\documentclass{article}

\PassOptionsToPackage{numbers}{natbib}

\usepackage[preprint]{neurips_2026}

\usepackage[utf8]{inputenc} 
\usepackage[T1]{fontenc}    
\usepackage{hyperref}       
\usepackage{url}            
\usepackage{booktabs}       
\usepackage{tabularray}     
\UseTblrLibrary{booktabs}
\usepackage{amsmath}        
\usepackage{amsfonts}       
\usepackage{nicefrac}       
\usepackage{microtype}      
\usepackage[table]{xcolor}  
\usepackage{graphicx}       
\usepackage{float}          

\title{Remember Smarter: Visual History Compressor and Hyperbolic Experience Space for Robotic Memory}

\author{%
  \begin{tabular}{c}
    Dai Zhou$^{1}$, 
    Jiexi Yan$^{1,\dagger}$, 
    Tong Li$^{1}$,
    Yuxuan Wang$^{1}$,
    Cheng Deng$^{1,\dagger}$
    \\[2pt]
    $^{1}$Xidian University 
    \\[2pt]
    $^{\dagger}$Corresponding author:
    \texttt{yanjiexi@xidian.edu.cn},
    \texttt{chdeng@mail.xidian.edu.cn}
  \end{tabular}
}

\begin{document}

\maketitle

\begin{figure}[H]
  \centering
  \includegraphics[width=\textwidth]{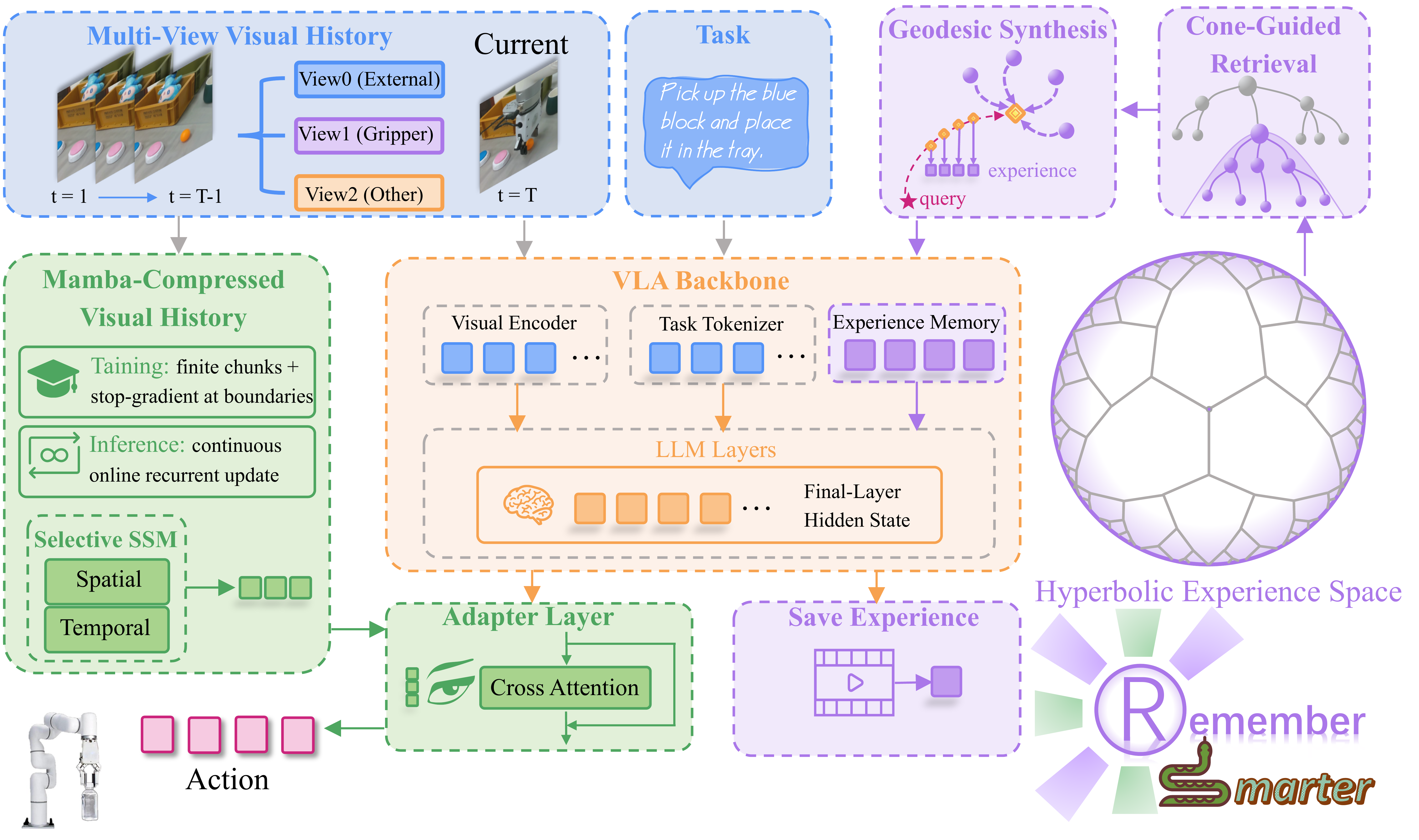}
  \caption{High-level concept of Remember Smarter. The fine-grained branch
  compresses multi-view visual patch history, while the coarse-grained branch
  retrieves structured hyperbolic experience. }
  \label{fig:overall-architecture}
\end{figure}

\begin{abstract}
  Long-horizon robot policies require compact access to recent observations and
  reusable experience without expanding the vision-language-action (VLA)
  context. We introduce Remember Smarter (RS), a plug-and-play module with
  complementary visual-history and hyperbolic experience-memory branches. Its
  visual branch compresses multi-view patch histories using bidirectional
  spatial Mamba and causal temporal Mamba, then exposes the resulting memory to
  action-facing hidden states through residual cross-attention while leaving the
  VLM visual-token stream unchanged. Its experience branch stores successful
  final-layer VLM states in a Poincare VAE space, organizes them hierarchically,
  and asynchronously converts retrieved experience into geodesic prompt tokens
  without blocking action inference. When adapted to $\pi_0$, RS increases total
  success on LIBERO-Plus from 53.6\% to 70.6\% and
  achieves substantial
  performance gains in real-robot experiments designed to evaluate memory
  retention and experience utilization.
\end{abstract}

\section{Introduction}
\label{sec:introduction}

Vision-language-action (VLA) models ground language instructions in visual
observations to predict robot actions
\citep{black2026pi0visionlanguageactionflowmodel,brohanRT2VisionLanguageActionModels2023,brohanRT1RoboticsTransformer2023,kimOpenVLAOpenSourceVisionLanguageAction2024b}.
Their memory, however, remains fragile on long-horizon tasks. Supplying more
historical frames to the backbone is costly and can be noisy: redundant
observations consume the token budget and can obscure the sparse state changes
needed for decision-making \citep{dai2026robomme,fang2025sam2act}. In addition,
a flat memory bank neither represents hierarchical abstractions explicitly nor
avoids global retrieval from a growing set of demonstrations
\citep{sridhar2025memer,mathieu2019pvae,ganea2018entailment}.

We propose \textbf{Remember Smarter (RS)}, a plug-and-play memory module that
combines visual-history compression with hyperbolic experience retrieval. For
fine-grained memory, RS compresses multi-view visual patch histories using
spatial and temporal Mamba blocks \citep{gu2023mamba,dao2024mamba2}.
Consecutive training fragments pass recurrent state across their boundaries
while stopping gradients every 16 frames; at inference, the recurrent state is
updated online. A residual cross-attention adapter lets action-facing hidden
states read the ordered history memory before the action projection without
modifying VLM visual tokens. For coarse-grained memory, RS collects only
successful training episodes and stores the final action-facing VLM
representation from each episode. A Poincare variational autoencoder
(P-VAE) supplies retrieval coordinates. An equal-radius auxiliary copy of the
leaves is optimized with a compact triplet objective inspired by Hyperbolic
Hierarchical Clustering (HypHC)~\citep{chami2020hyphc} and decoded into a binary
topology. A non-blocking beam search retrieves up to eight P-VAE
leaves, finds their lowest common internal ancestor, and uses that ancestor's
P-VAE-space prototype to guide four geodesic prompt tokens appended to the VLM
input.

Our contributions are summarized as follows:
\begin{itemize}
  \item We propose \textbf{Remember Smarter (RS)}, a plug-and-play memory
  architecture that augments VLA policies with complementary online
  visual-history memory and hierarchical experience memory without modifying
  the VLA backbone.

  \item We introduce an online visual-history compressor based on spatial and temporal Mamba, with state-passing truncated backpropagation through time (TBPTT) during training and fixed-state causal streaming at inference.

  \item We introduce a hierarchical hyperbolic experience space that combines
  P-VAE retrieval coordinates, binary-tree construction, entailment-cone
  routing, and geodesic experience-token generation.
\end{itemize}

\section{Related Works}
\label{sec:related-works}

\subsection{VLA Memory Mechanisms}

VLA policies combine visual perception, language grounding, and action
generation in a single model
\citep{brohanRT2VisionLanguageActionModels2023,kimOpenVLAOpenSourceVisionLanguageAction2024b,black2026pi0visionlanguageactionflowmodel}.
Recent lightweight designs improve the perception--action interface or reduce
backbone cost \citep{shukor2025smolvla,lin2025evo1,luo2026simvla,wang2026vlaadapter},
but long-horizon control also requires state that persists beyond one
observation. Existing memory-oriented policies use visual memory
\citep{fang2025sam2act,zhou2026aem}, phase or object-state modeling
\citep{fan2025longvla,chung2026objectmemory}, demonstration-derived prompts
\citep{li2025mapvla}, or multi-scale perceptual, cognitive, and language memory
\citep{shi2025memoryvla,shi2026memoryvlapp,torne2026mem}. RS
instead couples a fixed-state visual compressor with an asynchronously queried
experience space, assigning distinct computational roles to fine- and
coarse-grained memory.

Hierarchical organization has also been used to structure lifelong robot
experience \citep{baermann2025emv,nasir2019art} and multi-memory planning
\citep{zhou2025m2pa}. Related work studies hierarchical manipulation concepts
\citep{liu2025himacon} or action trees for manipulation world models
\citep{li2025manipdreamer}. In contrast, RS uses a decoded hierarchy only as a
retrieval index over successful policy representations, not as an explicit
symbolic task tree for action planning.

\subsection{State Space Models and Mamba}

Selective SSMs, including Mamba and its structured-state-space dual form, offer
linear-time recurrent sequence processing \citep{gu2023mamba,dao2024mamba2}.
Their long-context behavior nevertheless depends on state capacity,
initialization, and inter-layer information flow
\citep{chen2024stuffedmamba,wang2024longssm,he2025densemamba}; SSM reduction has
also been studied through balanced truncation \citep{ezoe2024balancedtruncation}.
For long video, STORM and
state-space hierarchical compression use temporal SSM encoders to preserve
relevant dynamics before reducing visual tokens
\citep{jiang2025storm,kim2026sshc}. RoboMamba applies Mamba directly within a
VLA policy \citep{liu2024robomamba}. RS uses Mamba differently: it retains the
pre-trained VLA backbone and compresses only the multi-view patch history into a
small state that the action-facing representations can read.

\subsection{Hyperbolic Geometry for Hierarchical Representation}

Poincare embeddings and Lorentz-model learning established hyperbolic geometry
as a useful inductive bias for latent hierarchies
\citep{nickel2017poincare,nickel2018lorentz}. Poincare VAEs, entailment cones,
and differentiable hyperbolic clustering respectively provide latent-variable,
directed-relation, and tree-construction tools
\citep{mathieu2019pvae,ganea2018entailment,chami2020hyphc}. RS uses P-VAE
coordinates for retrieval while optimizing a separate auxiliary copy only to
decode the hierarchy. This separation preserves the policy-facing retrieval space
and avoids treating the hierarchy-learning coordinates as online query features.

\section{Preliminary}
\label{preliminary}

\subsection{Problem statement}

We consider long-horizon manipulation with a pre-trained VLA backbone. At time
$t$, the robot receives a language instruction $L$, a multi-view image set
$I_t\in\mathbb{R}^{V\times H\times W\times3}$, and the preceding visual history
$\mathcal H_{1:t-1}=\{I_1,\ldots,I_{t-1}\}$. The visual-history compressor uses
this history as an auxiliary memory branch. The policy predicts
$a_t=\pi_\theta(I_t,L,M_t,\mathcal E)$, where $M_t$ is compact visual memory and
$\mathcal E$ is reusable experience from prior demonstrations. For manipulation,
$a_t$ is typically a 6-DoF end-effector pose, optionally augmented with a
gripper state for 7-DoF control.

\subsection{State Space Models}

State Space Models (SSMs) map a one-dimensional input sequence
$x(t)\in\mathbb{R}^{L}$ to an output sequence $y(t)\in\mathbb{R}^{L}$ through
a hidden state $h(t)\in\mathbb{R}^{N}$. A continuous SSM is defined by the
state matrix $\mathbf{A}$, input matrix $\mathbf{B}$, and output matrix
$\mathbf{C}$ \citep{gu2023mamba,dao2024mamba2}:
\begin{equation}
h'(t)=\mathbf{A}h(t)+\mathbf{B}x(t),\quad y(t)=\mathbf{C}h(t).
\end{equation}
Modern SSMs such as Mamba operate on discrete sequences. With timescale
parameter $\Delta$, zero-order-hold discretization converts $\mathbf A$ and
$\mathbf B$ into
\begin{equation}
\overline{\mathbf{A}}=\exp(\Delta \mathbf{A}),
\end{equation}
\begin{equation}
\overline{\mathbf{B}}=(\Delta \mathbf{A})^{-1}(\exp(\Delta \mathbf{A})-I)\cdot \Delta \mathbf{B},
\end{equation}
which yields the recurrent update
\begin{equation}
h_t=\overline{\mathbf{A}}h_{t-1}+\overline{\mathbf{B}}x_t,\quad y_t=\mathbf{C}h_t.
\end{equation}
Mamba instantiates this recurrent form as a selective SSM block. Rather than
using fixed transition and projection parameters for every token, it predicts
input-dependent parameters, including step size $\Delta_t$ and token-wise
projections $\mathbf B_t$ and $\mathbf C_t$. The resulting selective scan
processes a sequence in linear time, summarizes preceding tokens in a compact
state, and supports a parallel scan during training or recurrent state updates
at inference. For visual history, redundant frames can be absorbed into the
state while task-relevant changes still update the memory representation.

This recurrent view motivates state-passing truncated backpropagation through
time (TBPTT). During training, Mamba processes consecutive fragments of
multi-view visual patch tokens. The stop-gradient operator preserves the forward
state while limiting gradients to the current fragment. Our default setting uses
TBPTT of 16 and histories of at most 64 frames. We separately ablate history
lengths of 8, 16, 32, 64, and 128 frames and TBPTT truncation lengths of 4, 8,
16, 24, and 32 frames. At inference, the same constant-shape recurrent state
is updated online without storing the full visual history. Appendix~\ref{app:streaming-extrapolation}
gives the optimization details.

\subsection{Hyperbolic Space and Entailment Cones}

We briefly introduce the geometric concepts underlying hierarchical experience
representation, including hyperbolic space, the Poincare ball model, geodesics,
and entailment cones. This section focuses on their intuitive meanings and
relationships; the corresponding metrics, geometric operators, and formal
definitions are provided in Appendix~\ref{app:hyperbolic-formalization}.

\paragraph{Hyperbolic geometry.}
Hyperbolic space is a non-Euclidean space with constant negative curvature. Its
available volume grows exponentially with distance from the origin, resembling
the exponential increase in the number of nodes across successive levels of a
tree. This property makes hyperbolic space particularly suitable for
representing hierarchical data: general concepts can occupy a compact central
region, while increasingly specific concepts can spread over the much larger
outer regions. Compared with Euclidean embeddings, hyperbolic embeddings can
therefore preserve multi-level hierarchical relationships with relatively low
distortion \citep{nickel2017poincare,nickel2018lorentz}.

\paragraph{Poincare ball model.}
The Poincare ball is a conformal model of hyperbolic space in which all points
lie inside an open unit ball. Although the representation is bounded in
Euclidean coordinates, the boundary is infinitely far away under the
hyperbolic metric. Distances are increasingly stretched toward the boundary,
allowing the ball to accommodate a large number of well-separated
representations. Its radial organization also provides an intuitive
interpretation of abstraction: points near the origin generally represent
broad or abstract concepts, whereas points closer to the boundary represent
more specific concepts. Because the model is conformal, local angles are
preserved, which facilitates reasoning about both distance and direction.

\paragraph{Geodesics.}
A geodesic is the shortest path between two points under the hyperbolic metric.
It is the hyperbolic counterpart of a straight line in Euclidean space. In the
Poincare ball, geodesics are represented by diameters of the ball or circular
arcs that intersect its boundary orthogonally. Moving along a geodesic provides
a geometry-consistent interpolation between two representations. Points
sampled at different positions on the path capture progressively changing
levels of proximity to its endpoints without leaving the underlying
hyperbolic manifold.

\paragraph{Entailment cones.}
Entailment cones extend hyperbolic embeddings with directed hierarchical
relations. Each cone originates from an apex representing a relatively general
concept and covers a region associated with its more specific descendants. A
candidate representation is entailed by the apex when it lies inside this
directed region. Cone membership therefore encodes a directed
ancestor--descendant containment relation between the apex and the candidate,
rather than merely their undirected similarity. The cone aperture varies with the
location of its apex, enabling the geometry to express different levels of
generality and specificity while preserving transitive hierarchical
relationships \citep{ganea2018entailment}.

\section{Remember Smarter}
\label{sec:remember-smarter}

\subsection{Overview}
\label{sec:rs-overview}

Figure~\ref{fig:overall-architecture} summarizes RS, which augments a
pre-trained VLA policy with a visual-history compressor and a hierarchical
experience memory. The visual pathway summarizes changes in multi-view
observations; the experience pathway retrieves reusable patterns from
successful demonstrations. Together, they provide fine- and coarse-grained
memory without repeatedly extending the VLA context with raw history.

\begin{figure}[H]
  \centering
  \includegraphics[width=\textwidth]{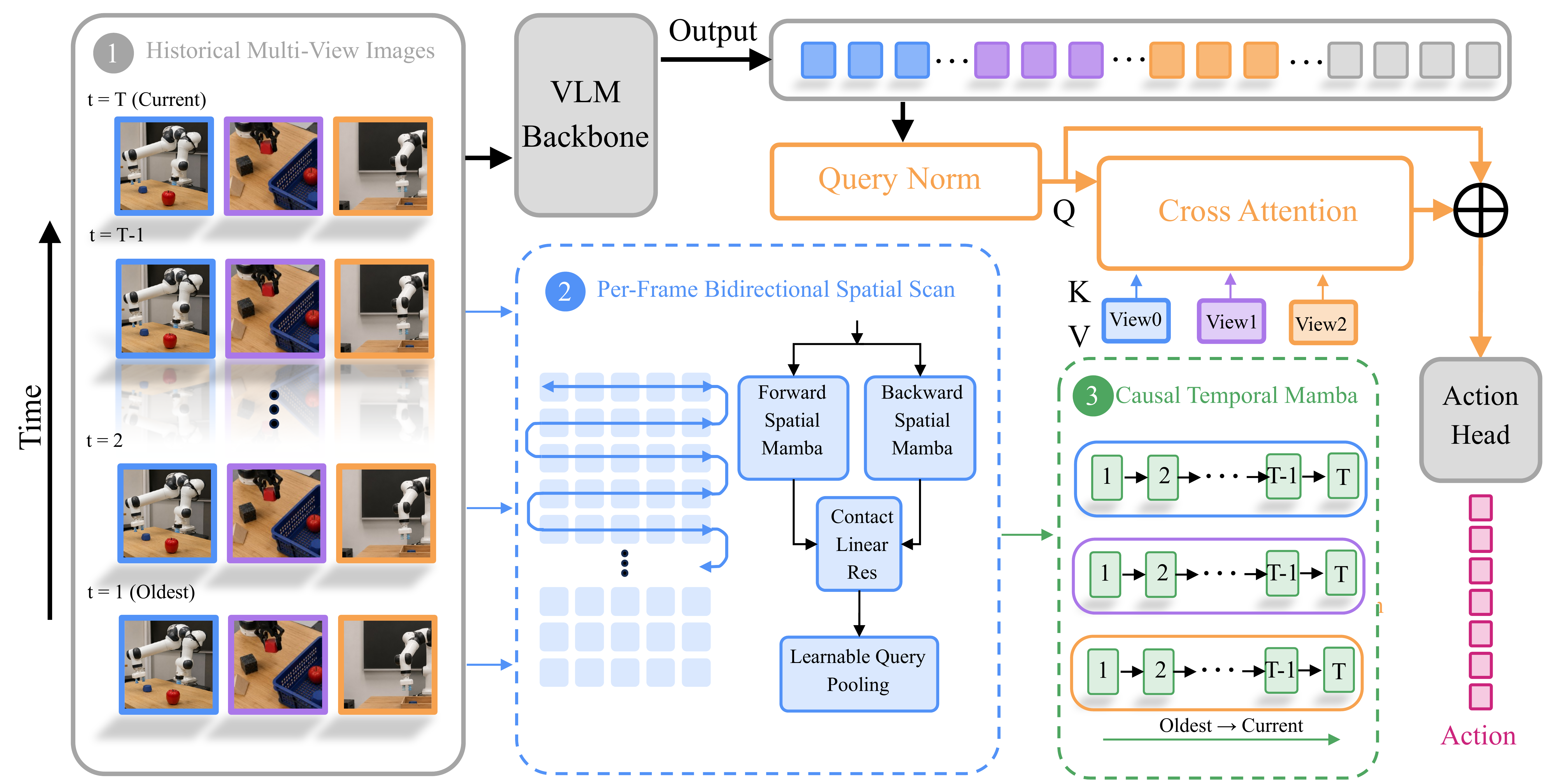}
  \caption{Mamba-compressed visual-history pipeline. Historical multi-view patch
  features are extracted by the shared visual encoder and mapped into a unified
  feature space by a lightweight adapter. Each frame is encoded through
  bidirectional spatial scans and learned-query pooling, after which
  causal temporal Mamba layers recurrently aggregate the per-view history. The
  resulting visual-history representations provide the keys and values for
  residual cross-attention, while the final hidden representations used for
  action prediction provide the queries. The enhanced hidden representations are
  then passed to the action head.}
  \label{fig:mamba-compressed-architecture}
\end{figure}

\subsection{Mamba-Compressed Visual History}
\label{sec:mamba-visual-history}

As illustrated in Figure~\ref{fig:mamba-compressed-architecture}, the
visual-history branch converts a multi-view image stream into a compact
recurrent memory for action prediction. It reuses the VLA visual encoder,
adapts its patch features to a shared latent width, applies bidirectional
spatial Mamba and learned-query pooling within each frame, and then aggregates
each view chronologically with causal temporal Mamba. A residual
cross-attention adapter lets the action-facing hidden states read this memory
without modifying the VLA visual-token stream. Appendix~\ref{app:mamba-implementation}
specifies the backbone paths, dimensions, masks, and training configuration.

\paragraph{Visual feature extraction and adaptation.}
Let $E_{\mathrm{vis}}$ denote the visual encoder already present in the VLA
policy. The current and historical images share this encoder, avoiding a second
visual backbone in the memory pathway. For view $v$ at time $\tau$, its patch
grid is adapted to the Mamba width $d$ by a lightweight module $A_\phi$:
\begin{equation}
  F_\tau^v=E_{\mathrm{vis}}(I_\tau^v),\qquad
  X_\tau^v=A_\phi(F_\tau^v)
  \in\mathbb R^{H_p\times W_p\times d}.
  \label{eq:visual-feature-adaptation}
\end{equation}
The adapter decouples the compressor from the native width and representation
format of a particular visual encoder, allowing the memory architecture to be
attached to VLA backbones with different visual feature dimensions.

\paragraph{Per-frame bidirectional serpentine encoding.}
For each frame and view, the adapted patch grid is linearized by a serpentine
permutation $\pi$: odd-numbered rows are traversed from left to
right, whereas even-numbered rows are traversed from right to left. This
alternating direction keeps consecutive patches spatially adjacent when the
scan crosses a row boundary. With $N_p=H_pW_p$, the resulting sequence is
\begin{equation}
  x_\tau^v =
  \left(X_\tau^v[\pi(1)],\ldots,X_\tau^v[\pi(N_p)]\right)
  \in\mathbb R^{N_p\times d}.
  \label{eq:patch-sequence}
\end{equation}
A forward spatial Mamba follows this sequence, whereas a backward spatial
Mamba follows its exact reverse. We reverse the latter output back to the
forward index system before fusion:
\begin{equation}
\begin{array}{rcl}
  \overrightarrow{Y}_\tau^v
    &=& \mathrm{Mamba}_{\mathrm{sp}}^{\rightarrow}(x_\tau^v), \\[2pt]
  \overleftarrow{Y}_\tau^v
    &=& \mathrm{Rev}\!\left(
       \mathrm{Mamba}_{\mathrm{sp}}^{\leftarrow}
       (\mathrm{Rev}(x_\tau^v))\right), \\[2pt]
  U_\tau^v
    &=& \tfrac{1}{2}\left(
       \overrightarrow{Y}_\tau^v+\overleftarrow{Y}_\tau^v
       \right).
\end{array}
  \label{eq:bidirectional-spatial-mamba}
\end{equation}
The two directions preserve aligned patch positions while providing
complementary spatial context. A learned query $q\in\mathbb R^d$ then pools
the fused sequence, and a learned projection $A_{\mathrm{pool}}$ produces a
frame-level representation:
\begin{equation}
  \alpha_\tau^v=\mathrm{softmax}_{N_p}\!\left(
  \frac{U_\tau^vq}{\sqrt{d}}\right),
  \qquad
  p_\tau^v=A_{\mathrm{pool}}\!\left((\alpha_\tau^v)^\top U_\tau^v\right).
  \label{eq:spatial-query-pooling}
\end{equation}

\paragraph{Causal temporal aggregation.}
For each view, the pooled frame representations are processed chronologically
by two causal temporal Mamba layers:
\begin{equation}
  (m_\tau^v,s_\tau^v) =
  \mathrm{Mamba}_{\mathrm{temp}}^{(2)}
  (p_\tau^v,s_{\tau-1}^v),
  \qquad \tau=1,\ldots,t.
  \label{eq:causal-temporal-mamba}
\end{equation}
Here, the superscript in $\mathrm{Mamba}_{\mathrm{temp}}^{(2)}$ denotes the
stacked two-layer temporal module. Thus, $m_\tau^v$ depends only on observations
available up to time $\tau$.
Temporal parameters are shared across views, while each view maintains its own
recurrent state $s_\tau^v$. The current multi-view memory has one ordered
representation for each view,
\begin{equation}
  M_t = \operatorname{stack}_{v\in\mathcal V}(m_t^v)
  \in\mathbb R^{B\times V\times d}.
  \label{eq:ordered-view-memory}
\end{equation}
This ordered history tensor serves as the memory source for the residual
cross-attention described below.
The state-update rule for unavailable views and the ordering convention are
specified in Appendix~\ref{app:mamba-implementation}.

\paragraph{Residual history cross-attention.}
Let $H_t\in\mathbb R^{B\times Q\times D_h}$ denote the final hidden
representations used for action prediction. RS applies one residual
cross-attention adapter,
\begin{equation}
\begin{aligned}
  Q_t &= H_tW_Q^{\mathrm{att}},\\
  K_t &= M_tW_K^{\mathrm{att}},\\
  V_t &= M_tW_V^{\mathrm{att}},\\
  \Delta H_t &= \mathrm{Attn}(Q_t,K_t,V_t)W_O^{\mathrm{att}},\\
  \widetilde H_t &= H_t+\gamma\,\Delta H_t,
\end{aligned}
  \label{eq:history-cross-attention}
\end{equation}
where $\mathrm{Attn}$ is multi-head scaled dot-product attention. The residual
scale is $\gamma=0.05$. The Mamba memory
supplies keys and values, action-facing hidden states supply queries, and the
original action head consumes the enhanced representation. Appendix~\ref{app:mamba-implementation}
details masking and backbone-specific insertion points.

\paragraph{Training and streaming inference.}
During training, each trajectory is divided into consecutive fragments. The
temporal state is passed forward within an episode, while the stop-gradient
operator detaches it at each TBPTT boundary:
\begin{equation}
  (Y_k^v,S_k^v) = \mathrm{Mamba}_{\mathrm{temp}}^{(2)}
  (P_k^v,\mathrm{stopgrad}(S_{k-1}^v)).
  \label{eq:visual-tbptt}
\end{equation}
In Eq.~\ref{eq:visual-tbptt}, $P_k^v$ is the sequence of pooled frame
representations in fragment $k$, $S_{k-1}^v$ is the incoming state, and $Y_k^v$
and $S_k^v$ are the output sequence and updated state. At inference, the same
fixed-shape recurrent state is updated one frame at a time, so prior images and
patch tokens need not be retained.

\subsection{Hyperbolic Experience Space}
\label{sec:hyperbolic-experience}

RS stores reusable experience in a hyperbolic hierarchy and retrieves it
asynchronously. Offline memory consists of P-VAE coordinates for successful
demonstrations and a HypHC-inspired binary topology over those coordinates. At
inference, a background worker queries this hierarchy and converts retrieved
experience into a fixed number of prompt tokens. Appendix~\ref{app:hyperbolic-formalization}
gives the formal geometry and implementation details.

\paragraph{Experience storage.}
After policy training, RS collects final action-facing hidden representations
$h_i$ from successful demonstrations. A Poincare variational autoencoder
(P-VAE)~\citep{mathieu2019pvae} maps each $h_i$ to posterior mean $z_i$ in the
Poincare ball. These means are retrieval leaves, and the P-VAE decoder maps
selected latent points back to the policy-representation space. Failed
demonstrations are not stored in experience memory.

\paragraph{Hierarchical organization.}
RS builds a separate binary hierarchy over the stored leaves with a compact
objective inspired by HypHC~\citep{chami2020hyphc}. Unlike a flat memory bank,
the hierarchy exposes nested experience groups for coarse-to-fine routing. This
stage optimizes an auxiliary copy of the leaf coordinates and therefore leaves
the P-VAE retrieval coordinates unchanged. After decoding the tree, each
internal prototype is initialized by the Karcher mean\citep{karcher1977riemannian,lou2020differentiating} of its descendant leaves,
which minimizes their aggregate squared hyperbolic distance. Entailment-cone
calibration then provides directed parent-to-descendant regions, complementing
the symmetric hyperbolic distance used for candidate ranking.

\begin{figure}[H]
  \centering
  \includegraphics[width=\linewidth]{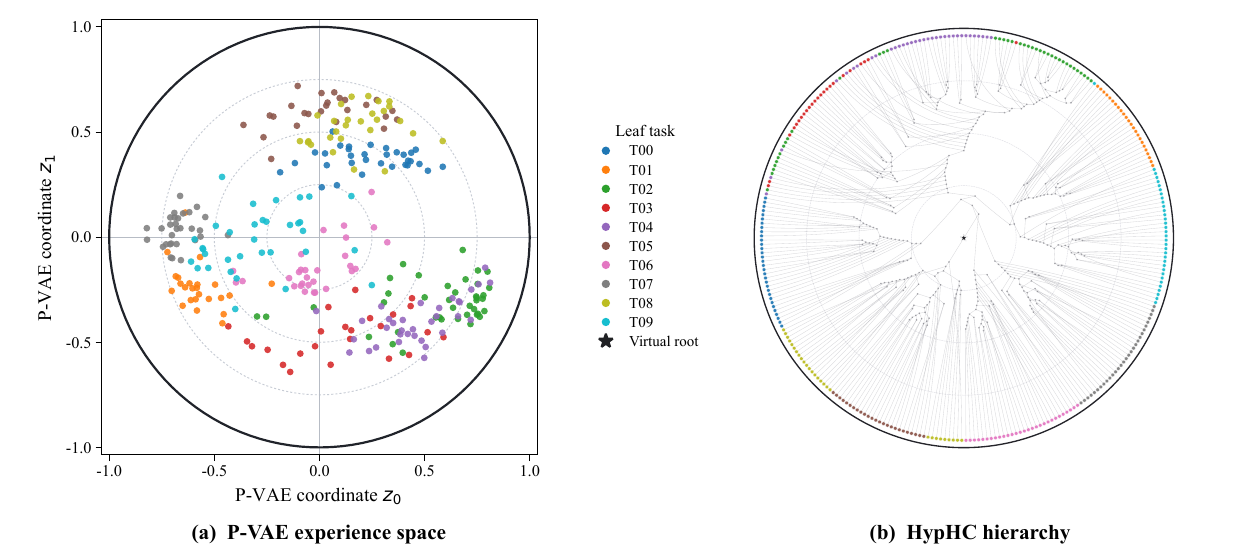}
  \caption{P-VAE experience space and HypHC hierarchy for the first 300
  successful $\pi_0$ inference records from
  \url{https://huggingface.co/datasets/physical-intelligence/libero}. Each
  colored point is a final-layer VLM hidden state encoded for visualization;
  color denotes its source task T00--T09. The black star in (b) marks the
  virtual root of the decoded HypHC hierarchy.}
  \label{fig:pvae-hyphc}
\end{figure}

\paragraph{Experience-space visualization.}
Figure~\ref{fig:pvae-hyphc} visualizes the P-VAE coordinates and corresponding
HypHC hierarchy for the first 300 successful $\pi_0$ inference records from the
released \texttt{physical-intelligence/libero} dataset. We encode each record's
final-layer VLM hidden state and plot the resulting leaves from its ten tasks
(T00--T09, listed in the appendix). The left panel shows P-VAE coordinates and
the right panel the decoded binary hierarchy. Colors identify the source task,
and the black star denotes the virtual root. This qualitative plot is not used
for online retrieval, which uses the full P-VAE coordinates and decoded tree.

\paragraph{Asynchronous hierarchical retrieval.}
At time step $t$, the current hidden representation is encoded as a query
$q_t$ in the same P-VAE space. A non-blocking worker traverses the binary tree,
using hyperbolic distance and the entailment cones of internal prototypes to
prioritize relevant subtrees before ranking candidate leaves. It returns at
most eight leaves, which bounds the subsequent composition cost. The action
loop never waits for this worker; queue management and reset handling are
specified in Appendix~\ref{app:hyperbolic-formalization}.

\paragraph{Geodesic experience-token generation.}
The retrieved leaves are summarized by their lowest common internal ancestor
$a_t$ in the binary hierarchy. Let $p_{a_t}$ denote the corresponding prototype
in the P-VAE space. RS samples four points along the geodesic from the query to
this shared prototype,
\begin{equation}
  u_j=\mathrm{Exp}_{q_t}\!\left(
  \tau_j\mathrm{Log}_{q_t}(p_{a_t})\right),\qquad
  e_j=D_\psi(u_j),\quad j=1,\ldots,4.
  \label{eq:geodesic-memory-tokens}
\end{equation}
where $0<\tau_1<\cdots<\tau_4<1$. Decoding these points produces four
experience tokens that progress from the current query toward the retrieved
shared abstraction. Sampling this geodesic supplies a fixed prompt budget that
is independent of the number of stored experiences. The four virtual
$\langle\mathrm{exp}\rangle$ slots are inserted immediately before the first
$\langle\mathrm{img}\rangle$ embedding. A learned [MEM\_QUERY] slot follows
the task prompt, and its final hidden representation forms the query for the
next retrieval. The original VLM self-attention fuses the added tokens with the
visual and language context; no separate experience-fusion module is used.

\section{Experiments}
\label{sec:experiments}

We evaluate RS only with $\pi_0$ in four settings: simulation on LIBERO-Plus,
real-robot experiments, component analysis of visual compression and
hyperbolic experience retrieval on standard LIBERO, and an analysis of how the
Mamba training-history window affects optimization and online compression.

\subsection{Simulation Benchmarks}

Following LIBERO-Plus \citep{fei2026liberoplus}, we evaluate robustness over
seven perturbation dimensions: camera viewpoint, robot initial state, language
instruction, lighting, background, sensor noise, and object layout. Both
$\pi_0$ and $\pi_0$+RS start from the official $\pi_0$--LIBERO checkpoint: the
foundation model is pre-trained on large-scale robot data and fine-tuned on the
original LIBERO demonstrations. RS is adapted only with those original LIBERO
demonstrations; it is not trained on LIBERO-Plus trajectories. We evaluate each
of the 10,030 LIBERO-Plus task instances once. Table~\ref{tab:liberoplus-results}
reports strict success rates over the full benchmark. RS raises total success
from 53.6\% to 70.6\%, a 16.7-point
increase. Its total is 2.5 points below the 73.1\% reported for MemoryVLA++ in
the zero-shot setting \citep{shi2026memoryvlapp}.

\begin{table*}[t]
  \centering
  \caption{Robustness on the full LIBERO-Plus benchmark. Entries are strict
  success rates (\%), and bold indicates the highest score in each reported
  column. Total is computed over all 10,030 tasks rather than averaging the seven
  dimension-level scores. Published baselines retain the settings reported in
  their original work; the MemoryVLA variants are evaluated in the zero-shot
  setting \citep{shi2026memoryvlapp}. The final row is the absolute change of
  $\pi_0$+RS relative to $\pi_0$.}
  \label{tab:liberoplus-results}
  \small
  \setlength{\tabcolsep}{5pt}
  \begin{tabular}{lcccccccc}
    \toprule
    Method & Camera & Robot & Language & Light & Background & Noise & Layout & Total \\
    \midrule
    MemoryVLA++ & 36.4 & \textbf{68.9} & \textbf{88.7} & \textbf{93.8} & 90.6 & 63.5 & 83.8 & \textbf{73.1} \\
    MemoryVLA & 42.7 & 44.9 & 84.4 & 92.8 & \textbf{95.0} & 62.1 & \textbf{84.7} & 70.2 \\
    OpenVLA-OFT & 56.4 & 31.9 & 79.5 & 88.7 & 93.3 & 75.8 & 74.2 & 69.6 \\
    $\pi_0$-Fast & \textbf{65.1} & 21.6 & 61.0 & 73.2 & 73.2 & 74.4 & 68.8 & 61.6 \\
    $\pi_0$ & 13.8 & 6.0 & 58.8 & 85.0 & 81.4 & 79.0 & 68.9 & 53.6 \\
    \midrule
    \rowcolor{blue!12}
    $\pi_0$+RS & 50.8 & 45.8 & 65.3 & 92.1 & 94.0 & \textbf{81.2} & 78.0 & 70.6 \\
    \textcolor{green!50!black}{$\Delta$ over $\pi_0$}
      & \textcolor{green!50!black}{$\uparrow$37.0}
      & \textcolor{green!50!black}{$\uparrow$39.8}
      & \textcolor{green!50!black}{$\uparrow$6.5}
      & \textcolor{green!50!black}{$\uparrow$7.1}
      & \textcolor{green!50!black}{$\uparrow$12.6}
      & \textcolor{green!50!black}{$\uparrow$2.2}
      & \textcolor{green!50!black}{$\uparrow$9.1}
      & \textcolor{green!50!black}{$\uparrow$16.9} \\
    \bottomrule
  \end{tabular}
\end{table*}

\subsection{Real-Robot Memory Tasks}

We evaluate $\pi_0$ and $\pi_0$+RS across seven real-robot task conditions;
their detailed definitions are given in Appendix~\ref{app:real-robot-tasks}.
The suite tests four memory-centric dimensions: repeated-action counting
(Press-2/4/6), ordered event sequences (Order), object--name alignment
(Chenchen), and compositional long-horizon execution (Chen.-H). Occluded
placement provides an additional perceptual stressor. We highlight the
compositional \emph{Chenchen-hard} (Chen.-H) task in the main text:
\emph{First, move Chenchen from the yellow box to the green box. Then press the
blue button once. Next, put the yellow ball into the green cup, and place the
green cup into the yellow box. Finally, press the pink button once.} The task
requires retaining object identity and recognizing the completion status of
multiple subgoals. RS is trained only on the simpler conditions and receives no
additional Chen.-H training.

Table~\ref{tab:real-robot-results} reports higher strict success for RS in six
of the seven conditions. On held-out Chen.-H, success rises from 20\% for
$\pi_0$ to 80\% for $\pi_0$+RS. The only decrease is on short-horizon Chen.-S
(64\% to 60\%); the largest increase is on Chen.-H (+60 points).

\begin{table}[t]
  \centering
  \caption{Real-robot strict success rate (\%). Press-$k$ denotes $k$
  repeated presses; Order denotes ordered button pressing; Occl. denotes
  occluded placement; Chen.-S/H denote the simple/hard Chenchen tasks.}
  \label{tab:real-robot-results}
  \small
  \setlength{\tabcolsep}{5pt}
  \begin{tabular}{lccccccc}
    \toprule
    Method & Press-2 & Press-4 & Press-6 & Order & Occl. & Chen.-S & Chen.-H \\
    \midrule
    $\pi_0$ & 64 & 20 & 0 & 20 & 72 & 64 & 20 \\
    \rowcolor{green!12}
    $\pi_0$+RS & \textbf{92} & \textbf{70} & \textbf{60} & \textbf{40} & \textbf{88} & 60 & \textbf{80} \\
    \bottomrule
  \end{tabular}
\end{table}

\begin{figure}[t]
  \centering
  \includegraphics[width=\linewidth]{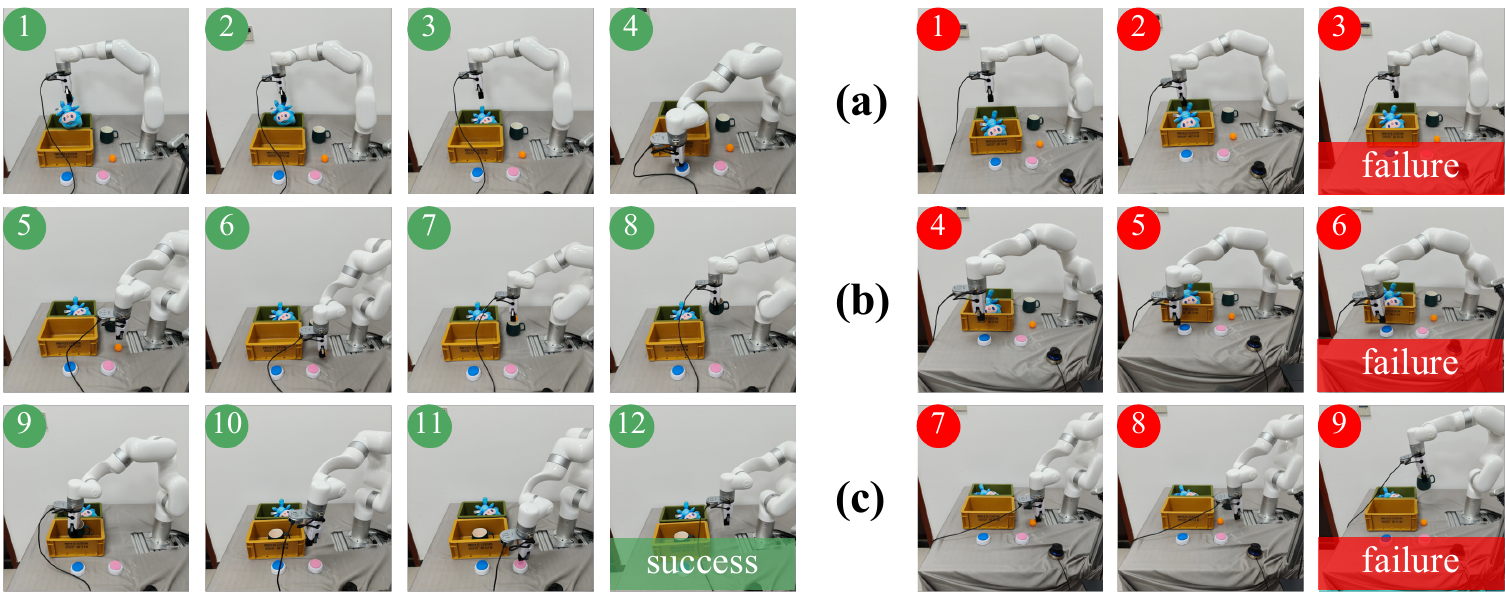}
  \caption{Representative real-robot rollouts on the compositional Chen.-H
  task. The green sequence shows a successful $\pi_0$+RS rollout. The red
  sequences show $\pi_0$ failures: (a) failure to lift Chenchen, (b) failure to
  press the required button, and (c) failure to place the ball in the cup.
  These examples are consistent with the difficulty of recognizing a failed
  subtask and recovering without access to successful experience.}
  \label{fig:real-robot-failures}
\end{figure}

\subsection{Ablations and Analysis}
\label{sec:ablations}

\begin{figure}[t]
  \centering
  \includegraphics[width=\linewidth]{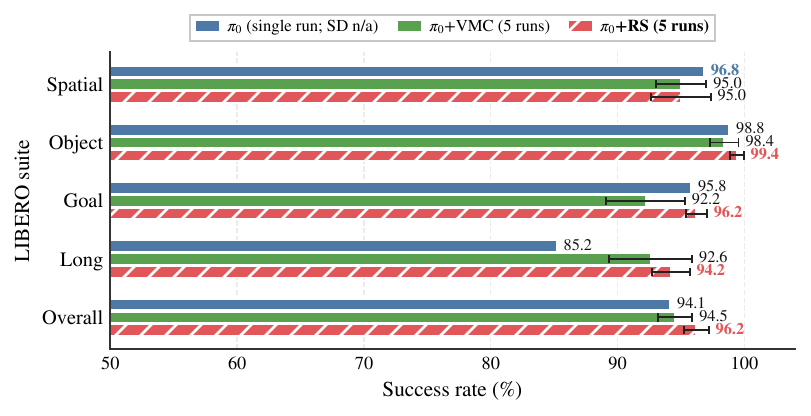}
  \caption{Component ablation on the four standard LIBERO suites. VMC is the
  visual Mamba compressor without the Poincare experience space. $\pi_0$ is a
  single run (standard deviation unavailable); $\pi_0$+VMC and $\pi_0$+RS use
  five runs, and error bars denote standard deviation.}
  \label{fig:libero-component-ablation}
\end{figure}

\paragraph{Component contributions.}
We compare the base $\pi_0$, $\pi_0$+VMC, and full $\pi_0$+RS on the four
standard LIBERO suites \citep{liu2023libero}. VMC contains only the visual
Mamba compressor; RS additionally enables the Poincare experience space. As
shown in Figure~\ref{fig:libero-component-ablation}, VMC raises Long success
from 85.2\% to 92.6\%, showing the value of visual-history compression. Adding
experience memory further raises Long to 94.2\% and the four-suite overall
score to 96.2\%, a 2.1-point improvement over $\pi_0$. The gain is concentrated
on the long-horizon suite: RS preserves comparable performance on the short
Spatial, Object, and Goal suites while yielding the largest improvement on
Long.

\begin{figure}[H]
  \centering
  \includegraphics[width=\linewidth]{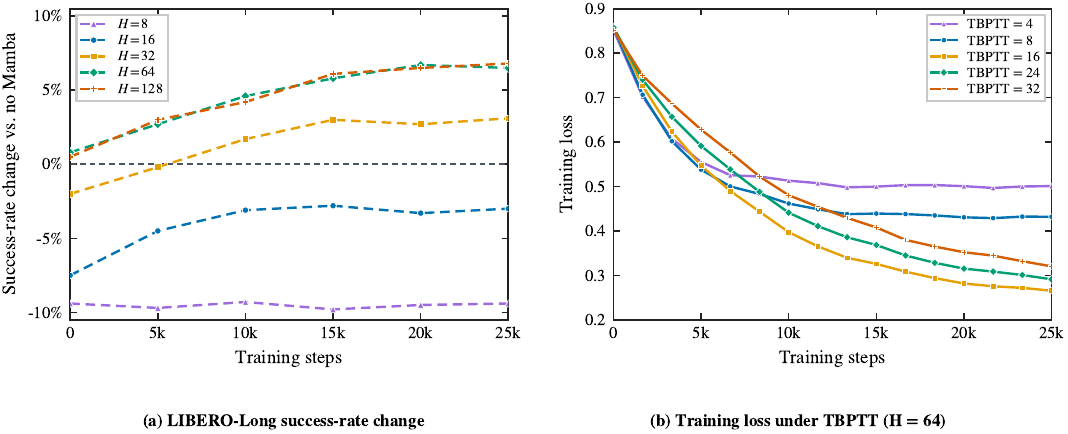}
  \caption{Effect of Mamba history length and TBPTT truncation. (a) Long
  success-rate change relative to a no-Mamba baseline, measured every 5k
  training steps. (b) Training loss for a fixed 64-frame history under different
  TBPTT truncation lengths.}
  \label{fig:mamba-window-ablation}
\end{figure}

\paragraph{History length and truncated backpropagation.}
Figure~\ref{fig:mamba-window-ablation}(a) measures LIBERO-Long success-rate
change relative to no Mamba every 5k training steps. Histories of 64 and 128
frames both reach roughly a 6\%--7\% improvement by 20k--25k steps, whereas
shorter histories remain weaker. Under this study, a 64-frame Mamba history
provides similar extrapolation to 128 frames without the longer context.
Figure~\ref{fig:mamba-window-ablation}(b) fixes the history at 64 frames and
varies the TBPTT gradient-truncation length. TBPTT=16 has the fastest sustained
loss reduction and the lowest loss at 25k steps; shorter windows plateau
earlier, while longer windows do not improve convergence.

\section{Conclusion}

We presented Remember Smarter, a plug-and-play memory framework that augments
VLA policies with fine-grained visual-history compression and hierarchical
experience retrieval. The visual branch reuses the VLA visual encoder, applies
bidirectional spatial and causal temporal Mamba, and maintains a compact
multi-view history in fixed-shape recurrent state. This memory supplies keys
and values to a residual cross-attention reader, while action-facing hidden
states supply queries and the original action head consumes the result.

The experience branch maps successful policy representations to P-VAE leaves
and organizes them with a separate HypHC-inspired binary topology. Karcher-mean
initialization and entailment-cone calibration yield internal prototypes for
routing. At inference, a non-blocking beam retrieves up to eight leaves; their
lowest common internal prototype guides four geodesic experience tokens. The
worker remains outside the action-critical path. On the selected LIBERO-Plus
subset, RS raises total success from 53.6\% to 70.3\%. It also achieves substantial performance gains in real-robot experiments designed to evaluate memory retention and experience utilization.
These results support the value of combining compact online visual history with
asynchronously retrieved successful experience for the evaluated long-horizon
settings.

\medskip

{
\small

\bibliographystyle{unsrt}
\bibliography{paper}
}


\appendix

\section{Mathematical Derivations}
\label{app:formula-derivations}

This appendix states the masked recurrence and residual-reader operators used
by RS. Let $\tilde s_\tau^v$ and $\tilde m_\tau^v$ be the state and output
proposed by the two-layer temporal Mamba module for view $v$. With availability
indicator $c_\tau^v\in\{0,1\}$, the implemented update is
\begin{equation}
  s_\tau^v=c_\tau^v\tilde s_\tau^v+(1-c_\tau^v)s_{\tau-1}^v,
  \qquad
  m_\tau^v=c_\tau^v\tilde m_\tau^v+(1-c_\tau^v)m_{\tau-1}^v.
  \label{eq:app-masked-recurrence}
\end{equation}
Thus, an unavailable camera neither changes its recurrent cache nor overwrites
its most recent memory token. For one attention head with valid-view mask
$c_t^v$, the history reader computes
\begin{equation}
  a_{ij}=\frac{\exp(q_i^\top k_j/\sqrt{d_h})c_t^{v_j}}
  {\sum_{r=1}^{V}\exp(q_i^\top k_r/\sqrt{d_h})c_t^{v_r}},
  \qquad
  \widetilde h_i=h_i+\gamma W_O\sum_{j=1}^{V}a_{ij}v_j,
  \label{eq:app-masked-history-reader}
\end{equation}
where all attention weights are defined as zero when no view is valid. The
multi-head implementation applies Eq.~\ref{eq:app-masked-history-reader} to
each head and concatenates the resulting contexts. Finally, the four retrieved
experience tokens are points on the Poincare geodesic
\begin{equation}
  \Gamma_{q,p}(\tau)=\operatorname{Exp}_{q}
  \bigl(\tau\operatorname{Log}_{q}(p)\bigr),
  \qquad \tau\in\{0.45,0.55,0.65,0.75\},
  \label{eq:app-geodesic-derivation}
\end{equation}
from the current query $q$ to the prototype $p$ of the retrieved lowest common
ancestor. These points are decoded independently before entering the original
VLM self-attention stack.

\section{Detailed Framework Description}
\label{app:framework-details}

\subsection{Implementation Details of Mamba-Compressed Visual History}
\label{app:mamba-implementation}

\paragraph{Backbone feature paths and adapters.}
The history pathway reuses the frozen $\pi_0$ visual encoder
\citep{black2026pi0visionlanguageactionflowmodel} through the shared openpi
policy interface \citep{physicalintelligenceopenpi}. The frozen SigLIP
\citep{zhai2023siglip} forward pass returns $16\times16$ patches through
\texttt{pre\_logits\_2d}. After stop-gradient, an independently trained
$\mathrm{LayerNorm}(D_{\mathrm{SigLIP}})+\mathrm{Linear}(D_{\mathrm{SigLIP}},256)$
adapter maps these patches to the compressor width. Original VLM image tokens
are unchanged. No PCA or auxiliary distillation objective is used. Training
caches contain complete episode patch sequences and view masks, but never Mamba
memories or recurrent states.

\paragraph{Compressor configuration and view masking.}
The compressor width is $d=256$, with state size 16, convolution width 4,
expansion factor 2, and head width 64. One Mamba layer operates in each spatial
direction; aligned outputs are averaged, pooled by one learned query, and
passed through a learned projection. Two causal temporal Mamba layers then
update per-view histories. Temporal parameters are shared across the ordered
multi-view stream, whereas convolution and SSM states remain separate for each
view. For availability mask
$c_\tau^v\in\{0,1\}$, the temporal update is
\begin{equation}
  (m_\tau^v,s_\tau^v)=
  \begin{cases}
  \mathrm{Mamba}_{\mathrm{temp}}^{(2)}(p_\tau^v,s_{\tau-1}^v), & c_\tau^v=1,\\
  (m_{\tau-1}^v,s_{\tau-1}^v), & c_\tau^v=0,
  \end{cases}
  \label{eq:app-view-mask-update}
\end{equation}
so an unavailable view changes neither its recurrent state nor its latest
memory representation.

\paragraph{Residual reader and backbone seams.}
The history reader uses four attention heads of width 64 and residual scale
$\gamma=0.05$. Memory masks remove unavailable views from the attention
logits, and an item with no valid memory receives an exactly zero residual
update. Disabling visual history bypasses the reader and preserves the original
action path. The abstract
$H_t$ in Eq.~\ref{eq:history-cross-attention} denotes the final hidden
representations used for action prediction. For $\pi_0$, it is instantiated by
\texttt{suffix\_out} before \texttt{action\_out\_proj} at every flow-matching
step. This provides a single residual history-reading interface while retaining
the $\pi_0$ policy parameters and independently trained RS modules.

\paragraph{Online state contract.}
At inference, each sampled frame performs one spatial encoding and one
mask-conditional temporal update for every available view. Episode reset clears
convolution caches, SSM states, latest view memories, and the sampling timer, so
no state crosses episode boundaries. Recurrent-state shape is independent of
episode length. Appendix~\ref{app:training-configuration} gives TBPTT and
optimization details.

\subsection{Formalization of the Hyperbolic Experience Space}
\label{app:hyperbolic-formalization}

\paragraph{Poincare geometry and numerical convention.}
We use curvature magnitude $c=1$ and the open ball $\mathbb B^d$. For floating
point tensors, $\epsilon=10^{-7}$ for float32/bfloat16/float16 and $10^{-12}$
for float64. The implementation clamps \texttt{artanh} inputs to
$[-1+\epsilon,1-\epsilon]$ and projects points to radius $1-10\epsilon$. With
$\lambda_x=2/(1-\|x\|^2)$ and Mobius addition $\oplus$, the common operations are
\begin{equation}
 d_{\mathbb B}(x,y)=2\,\mathrm{artanh}(\|-x\oplus y\|),
 \label{eq:app-poincare-distance}
\end{equation}
\begin{align}
 \mathrm{Exp}_x(v)&=x\oplus
 \left(\tanh\!\frac{\lambda_x\|v\|}{2}\frac{v}{\|v\|}\right),
 \qquad
 \mathrm{Log}_x(y)=\frac{2\,\mathrm{artanh}(\|-x\oplus y\|)}
 {\lambda_x\|-x\oplus y\|}(-x\oplus y).
 \label{eq:app-general-maps}
\end{align}
Norms and denominators are clamped by $\epsilon$, with continuous zero limits.
A Karcher solve starts from the projected weighted Euclidean mean and runs 12
updates $m\leftarrow\mathrm{Exp}_m(0.5\sum_iw_i\mathrm{Log}_m z_i)$.

\paragraph{P-VAE leaves and query.}
For every eligible final-layer hidden state
$h_i\in\mathbb R^{D_{\mathrm{VLM}}}$, the wrapped-normal P-VAE uses
\begin{equation}
 z_i=\mu_i=\mathrm{Exp}_0(0.25\tanh f_\mu(h_i)),\qquad
 \sigma_i=\mathrm{softplus}(\mathrm{clamp}(f_\sigma(h_i),-8,4))+10^{-4},
 \label{eq:app-pvae-location-scale}
\end{equation}
and optimizes reconstruction plus $\beta$ KL. Its decoder is an MLP on
$\mathrm{Log}_0(\Pi(z))$. Sampling is used in training, whereas retrieval always
uses projected posterior locations. The successful-episode filter is applied
before the immutable snapshot is built. The online [MEM\_QUERY] hidden follows
the same $z=\mu_\phi(h)$ path to obtain $q_t$; it has no additional head or cone
semantics.

\paragraph{Compact HypHC-inspired triples and tree decode.}
Let $Z=(z_1,\ldots,z_N)$ be P-VAE leaves and let $\rho=0.75$. An independent
auxiliary copy is initialized as
\begin{equation}
  y_i=\Pi\!\left(
  \rho\frac{z_i}{\max(\|z_i\|,10^{-6})}\right),
  \label{eq:app-routing-coordinate-initialization}
\end{equation}
where $\Pi$ is the Poincare-ball projection. Gradients through this phase do
not modify $Z$. We form
$s_{ij}=\exp[-d_{\mathbb B}^2(z_i,z_j)/(2b^2)]$, where $b$ is the median
nonzero pairwise P-VAE distance, and sample distinct triples $(a,b,c)$. For the
three pairs $\mathcal P=\{(a,b),(a,c),(b,c)\}$, the implemented compact loss is
\begin{equation}
  w_{ij}=\frac{\exp[-d_{\mathbb B}(y_i,y_j)]}
  {\sum_{(r,s)\in\mathcal P}\exp[-d_{\mathbb B}(y_r,y_s)]},
  \qquad
  \mathcal L_{\mathrm{tri}}=
  \mathbb E\!\left[\left(\sum_{(i,j)\in\mathcal P}s_{ij}
  -\sum_{(i,j)\in\mathcal P}w_{ij}s_{ij}\right)^2\right].
  \label{eq:app-hyphc-inspired-triplet}
\end{equation}
After each update, $y_i$ is reprojected to radius $\rho$, subject to the
numerical floor in Eq.~\ref{eq:app-routing-coordinate-initialization}. This is a
differentiable triplet relaxation inspired by HypHC, rather than the full HypHC
training system.

All leaf pairs are sorted by descending dot product, with stable leaf-ID ties.
A Union--Find single-link pass merges disconnected components until the
deterministic binary tree has $2N-1$ nodes. The decoded root serves as a virtual
router. For retrieval-space initialization, each internal prototype is
recomputed as the Karcher mean of its descendant P-VAE leaves $z_i$;
equal-radius $y_i$ values remain auxiliary.

\paragraph{Topology-only cone calibration.}
For $K=0.1$, a non-root internal prototype $p$ has half-aperture and exclusion
radius
\begin{equation}
 \psi(p)=\arcsin\!\left(K\frac{1-\|p\|^2}{\|p\|}\right),
 \qquad r_{\min}=\frac{2K}{1+\sqrt{1+4K^2}}.
 \label{eq:app-cone-domain}
\end{equation}
The angle is computed between $-\mathrm{Log}_p(0)$ and
$\mathrm{Log}_p(x)$, with cosine clamped to
$[-1+10^{-7},1-10^{-7}]$, and
$E_{\mathrm{cone}}(p,x)=[\angle_p(x)-\psi(p)]_+$. If
$\|p\|<r_{\min}$, this energy is marked invalid and routing uses distance only.

Topology supplies positives from non-root internal-to-child edges and
internal-to-descendant-leaf closure. For each positive parent, outside-subtree
leaves with the lowest initial cone energy supply hard negatives. With an
adaptive radial margin $m_{px}$, calibration minimizes
\begin{equation}
\begin{aligned}
 \mathcal L_{\mathrm{cal}}={}&
 \lambda_+\mathbb E_{(p,x)\in\mathcal P}E_{\mathrm{cone}}(p,x)
 +\lambda_-\mathbb E_{(p,n)\in\mathcal N}[\gamma-E_{\mathrm{cone}}(p,n)]_+ \\
 &+\lambda_r\mathbb E_{(p,x)\in\mathcal P}[\|p\|+m_{px}-\|x\|]_+
 +\lambda_a\mathbb E_p d_{\mathbb B}^2(p,p^{(0)}).
\end{aligned}
\label{eq:app-topology-cone-calibration}
\end{equation}
Relations are generated solely from the decoded topology and never read task,
subtask, or action labels. Optimization variables are tangent coordinates for
non-root internal prototypes only. P-VAE leaves and the virtual root are copied
back exactly after optimization.

\paragraph{Beam and exact retrieval.}
For a non-root internal node $v$, the routing score is
\begin{equation}
  s_{\mathrm{node}}(v\mid q_t)=
  0.25\,d_{\mathbb B}(p_v,q_t)+E_{\mathrm{cone}}(p_v,q_t).
  \label{eq:node-score}
\end{equation}
The approximate search begins at the virtual root's children, retains the eight
lowest-scoring internal nodes at each level, and expands their children. Visited
leaves accumulate until the usable pool reaches $\max(8,\lceil2\cdot8\rceil)=16$.
A visited leaf $\ell$ is ranked by
\begin{equation}
  s_{\mathrm{leaf}}(\ell\mid q_t)=d_{\mathbb B}(q_t,z_\ell)
  +\frac{1}{|\mathcal A_\ell^{\mathrm{valid}}|}
   \sum_{v\in\mathcal A_\ell^{\mathrm{valid}}}
   E_{\mathrm{cone}}(p_v,q_t),
  \label{eq:final-leaf-score}
\end{equation}
where the second term is zero if no ancestor has a valid cone. After excluding
records whose episode ID equals the query episode, the best eight leaves are
returned. If exclusions
leave fewer than the required number, an explicit exact safety completion
ranks all leaves.

The exact oracle batches all non-root internal cone energies, propagates
ancestor sums and valid counts with one root-to-leaf DFS, batches all leaf
distances, and performs the same final top-8 sort. Its per-query cost is
$\mathcal O(N)$. The beam records visited and pruned nodes and its fallback
reason, but has no sublinear worst-case guarantee and may also cost
$\mathcal O(N)$.

\paragraph{Composer and fixed output.}
The composer finds the lowest common internal ancestor $a_t$ of the returned
top-8 leaves in the decoded topology and obtains its calibrated P-VAE-space
prototype $p_{a_t}$. For zero-based slot $j$, the implementation uses
\begin{equation}
  (\tau_0,\tau_1,\tau_2,\tau_3)=(0.45,0.55,0.65,0.75),
 \qquad u_j=\mathrm{Exp}_{q_t}(\tau_j\mathrm{Log}_{q_t}p_{a_t}).
 \label{eq:app-composer-schedule}
\end{equation}
Decoding $u_0,\ldots,u_3$ yields exactly $4\times D_{\mathrm{VLM}}$ values.
Empty candidates return four zero decoder-width slots to the non-blocking caller
but are classified as empty retrieval and never published by the asynchronous
worker.

\paragraph{Asynchronous validity.}
One global capacity-one request queue uses latest-wins submission. Every
submission replaces any queued request, increments a sequence number, and
records the latest tuple $(g,s,e,t)$ of generation, sequence, episode, and
step. A completed worker result is published only when its tuple still matches
the latest request and elapsed wall time is within the configured timeout. Empty
results, exceptions, and post-completion timeouts increment telemetry and are
discarded. The action loop never waits. Poll validity is
\begin{equation}
 \mathrm{valid}(\mathcal R_\tau,t)=
 \mathbb I[e_\tau=e_t]\,\mathbb I[g_\tau=g_t]\,
 \mathbb I[1\leq t-\tau\leq2].
 \label{eq:app-async-validity}
\end{equation}
Thus, an age-zero result cannot affect the current action step. Reset increments
the generation, clears episode results and step state, and drains the queue.
Search also excludes the query episode before ranking.

\paragraph{Prompt attention contract.}
For each batch row, four projected $\langle\mathrm{exp}\rangle$ embeddings are
inserted immediately before the first real image embedding, and one learned
[MEM\_QUERY] embedding is inserted at the end of the task prompt. Position IDs
are recomputed from the valid-token mask, and decoder-only adapters use the
corresponding causal or prefix-block attention edges. The existing VLM processes
this augmented prompt with its original self-attention. From the final hidden
sequence, [MEM\_QUERY] is detached and submitted for use only at $t+1$ or
later; original
token positions are gathered back before the action expert.

No production experience cross-attention or second residual gate exists.
With \texttt{enabled=false}, no virtual tokens are inserted and inputs,
attention mask, and original action path are returned unchanged. With experience
enabled but no memory yet, masked empty experience slots and a valid
[MEM\_QUERY] are still inserted so that a future query can be produced.

\section{Experimental Configuration and Additional Results}
\label{app:experimental-configuration}

\subsection{Complete Evaluation Protocol}

For every paired comparison, evaluation instances and initial states are fixed
before either policy is run. LIBERO-Plus uses all 10,030 single-perturbation
instances described in Section~\ref{sec:experiments}. The
reported primary setting trains only on original LIBERO; the released
LIBERO-Plus training set is not used. The standard LIBERO component study uses
each suite's demonstrations for the corresponding Spatial, Object, Goal, or
Long evaluation.

The frozen-backbone study trains the SigLIP dimension adapter, bidirectional
spatial Mamba blocks, learned-query pooling and projection, two temporal Mamba
layers, the residual history reader, and the implemented experience components:
P-VAE, independent topology-learning coordinates, internal-prototype
calibration, prompt projection, and [MEM\_QUERY] embedding. It includes no
query-radius head, experience cross-attention, or second residual gate.

\subsection{Mamba Training-Window and Fixed-State Streaming Evaluation}
\label{app:streaming-extrapolation}

For the history-length study, we train VMC with Mamba histories of
$H\in\{8,16,32,64,128\}$ and evaluate the LIBERO-Long success-rate change
relative to a no-Mamba baseline every 5k steps. The TBPTT study fixes $H=64$
and varies the backpropagation truncation length over
$\{4,8,16,24,32\}$. Training loss is logged throughout optimization. The
main paper reports the resulting success-rate and loss curves; the supplement
contains additional optimization diagnostics.

\subsection{Component Ablation Details}
\label{app:experiment-details}

The component ablation uses only $\pi_0$ and compares the base policy,
$\pi_0$+VMC, and full $\pi_0$+RS. Each variant is trained and evaluated under
the standard protocol for LIBERO-Spatial, LIBERO-Object, LIBERO-Goal, and
LIBERO-Long. No variant changes the demonstrations, data order, optimization
budget, checkpoint-selection rule, action parameterization, or evaluation
instances within a suite. The grouped plot reports each suite and their
unweighted Overall mean. The base $\pi_0$ result is a single run; VMC and RS
report five runs, with standard deviation shown in the main-paper figure.

\subsection{P-VAE and HypHC Visualization Data}

Figure~\ref{fig:pvae-hyphc} uses the first 300 successful $\pi_0$ inference records from
\url{https://huggingface.co/datasets/physical-intelligence/libero}. The records
span the following ten tasks:

\begin{tabular}{@{}lp{0.82\linewidth}@{}}
\toprule
Task & Instruction \\
\midrule
T00 & Put the white mug on the left plate and put the yellow and white mug on the right plate. \\
T01 & Put the white mug on the plate and put the chocolate pudding to the right of the plate. \\
T02 & Put the yellow and white mug in the microwave and close it. \\
T03 & Turn on the stove and put the moka pot on it. \\
T04 & Put both the alphabet soup and the cream cheese box in the basket. \\
T05 & Put both the alphabet soup and the tomato sauce in the basket. \\
T06 & Put both moka pots on the stove. \\
T07 & Put both the cream cheese box and the butter in the basket. \\
T08 & Put the black bowl in the bottom drawer of the cabinet and close it. \\
T09 & Pick up the book and place it in the back compartment of the caddy. \\
\bottomrule
\end{tabular}

\subsection{Real-Robot Task Definitions and Success Criteria}
\label{app:real-robot-tasks}

The real-robot suite probes four memory-centric requirements: counting repeated
events, preserving an ordered action sequence, grounding a persistent object
name, and composing these abilities over a long horizon. Occluded placement is
an additional perceptual stressor. The task instructions are as follows:
\begin{itemize}
  \item \textbf{Press-2}: \emph{Press the purple button twice.} Press-4 and
  Press-6 use the same button with the requested count changed to four and six.
  \item \textbf{Order (Press-hard)}: \emph{First, press the blue button once.
  Then press the purple button twice. Finally, press the pink button three
  times.}
  \item \textbf{Occl.}: \emph{First, put the yellow ball into the green cup,
  then place the green cup into the red box.}
  \item \textbf{Chen.-S}: \emph{Move Chenchen from the yellow box to the green
  box.} Here, \emph{Chenchen} always denotes the designated plush toy rather
  than a distractor.
  \item \textbf{Chen.-H}: \emph{First, move Chenchen from the yellow box to
  the green box. Then press the blue button once. Next, put the yellow ball
  into the green cup, and place the green cup into the yellow box. Finally,
  press the pink button once.}
\end{itemize}
Strict success requires completion of every requested subgoal in the specified
order. Any wrong object, wrong order, extra button press, safety stop, timeout,
or human intervention is a failure.

\paragraph{Training and generalization conditions.}
Training uses Press-2, Ordered Button Pressing, Occluded Placement, and the
Chenchen-simple condition. Press-4, Press-6, and Chenchen-hard are evaluated
without additional task-specific training.

Each physical condition is evaluated over 10 or 25 trials. In addition to
strict success rate, logging records button-count error, the first sequence
error, redundant actions after hidden placement, completed-subgoal fraction,
and the first failure stage.

\begin{figure}[H]
  \centering
  \includegraphics[width=\linewidth]{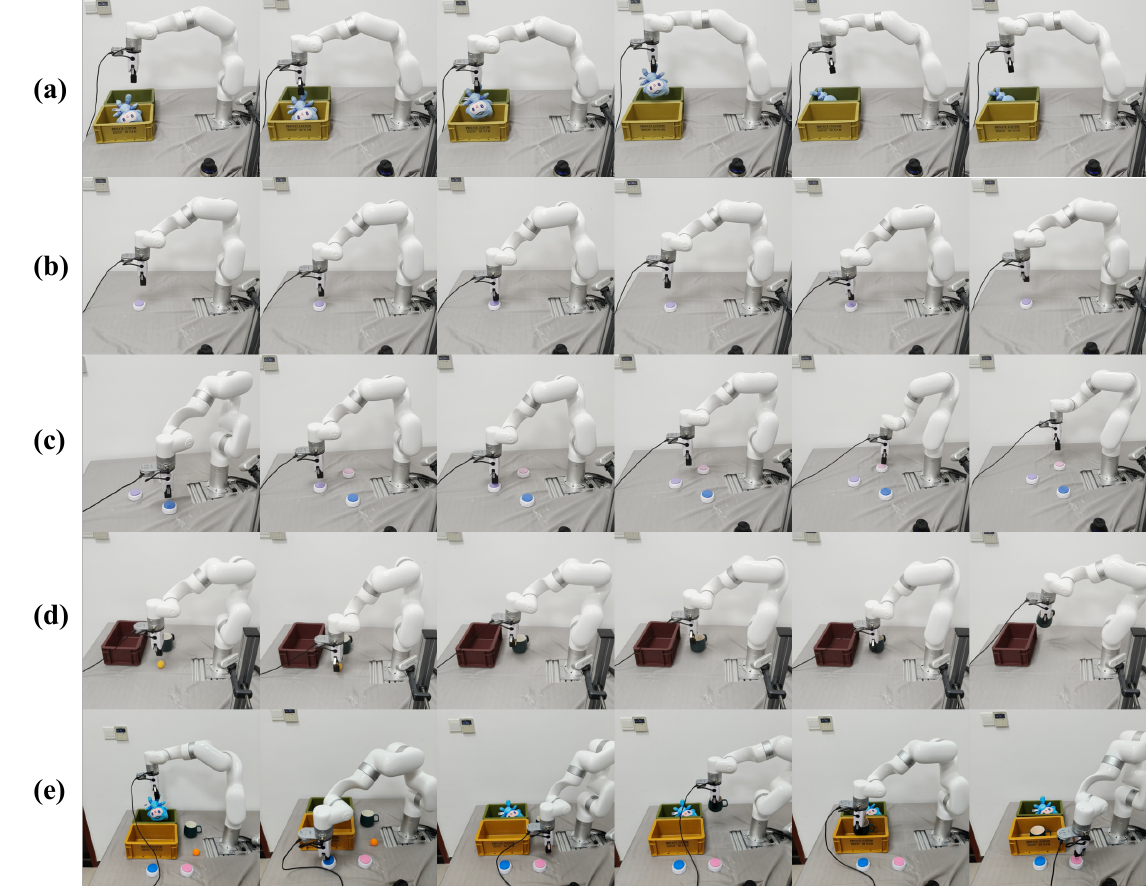}
  \caption{Representative successful keyframe sequences for the real-robot
  task families: (a) Chen.-S name grounding, (b) Press-2 counting, (c) Order
  sequence following, (d) occluded placement, and (e) Chen.-H compositional
  long-horizon execution.}
  \label{fig:real-robot-task-overview}
\end{figure}

\section{Training Configuration}
\label{app:training-configuration}

RS is trained through the pinned openpi $\pi_0$ interface. The released
$\pi_0$ policy checkpoint initializes the policy, while the visual compressor
starts from the seed-0 \texttt{mamba\_base}; no additional standalone Mamba
pretraining is used. Cached inputs contain frozen SigLIP
\texttt{pre\_logits\_2d} patch features and view masks, rather than cached
recurrent activations. Each example therefore reconstructs its causal history
using the current RS parameters.

\begin{table}[H]
  \centering
  \caption{Default $\pi_0$+RS training configuration.}
  \label{tab:training-configuration}
  \small
  \setlength{\tabcolsep}{5pt}
  \begin{tabular}{@{}p{0.32\linewidth}p{0.58\linewidth}@{}}
    \toprule
    Item & Setting \\
    \midrule
    Visual history & 3 views, 1 FPS, 64 frames; TBPTT window 16 \\
    VMC & $d=256$, state 16, convolution 4, expansion 2, head width 64 \\
    Precision / parallelism & bf16 activations; fp32 parameters and gradients; 4-device FSDP \\
    Optimizer & Grouped AdamW; weight decay $10^{-4}$; gradient clip 1; EMA 0.999 \\
    Batch and checkpoints & Global batch 32; checkpoint every 2,500 steps \\
    Stage A & 5k steps; RS adapter, VMC, and history reader trainable \\
    Stage B & 25k steps; RS modules and action-side parameters trainable; PaliGemma prefix frozen \\
    Learning rates & Stage A: $2\times10^{-5}$ (Mamba), $10^{-4}$ (adapter/reader); \\
    & Stage B: $10^{-5}$ (Mamba/adapter), $5\times10^{-5}$ (reader/action) \\
    \bottomrule
  \end{tabular}
\end{table}

Each nonzero learning-rate group uses a 100-step warm-up followed by cosine
decay to one tenth of its peak rate. The default evaluation uses a 64-frame
history. The history-length and TBPTT studies reported in
Figure~\ref{fig:mamba-window-ablation} vary those values separately; they are
not part of the default training setting.



\end{document}